\documentclass[9pt,conference]{IEEEtran}
\IEEEoverridecommandlockouts

\usepackage{cite}
\usepackage{amsmath,amssymb,amsfonts}
\usepackage{algorithmic}
\usepackage{graphicx}
\usepackage{textcomp}
\usepackage{xcolor}
\usepackage{color,amssymb,multirow}
\usepackage[flushleft]{threeparttable}
\usepackage{booktabs,subfigure,kotex}
\usepackage{url}
\usepackage{FiraMono}
\usepackage{enumitem}
\usepackage{xcolor}
\usepackage{fontawesome}
\usepackage{listings}

\makeatletter
\newcommand*{\rom}[1]{\expandafter\@slowromancap\romannumeral #1@}
\makeatother
\makeatletter
\DeclareRobustCommand{\textsupsub}[2]{{%
  \m@th\ensuremath{%
    ^{\mbox{\fontsize\sf@size\z@#1}}%
    _{\mbox{\fontsize\sf@size\z@#2}}%
  }%
}}
\makeatother

\newcommand{\sy}[1]{\textcolor{black}{#1}}

\definecolor{gt}{RGB}{0, 0, 0}
\newcommand{\gt}[1]{\textcolor{gt}{#1}}

\definecolor{nam}{RGB}{0, 0, 0}
\newcommand{\nam}[1]{\textcolor{nam}{#1}}

\def\BibTeX{{\rm B\kern-.05em{\sc i\kern-.025em b}\kern-.08em
    T\kern-.1667em\lower.7ex\hbox{E}\kern-.125emX}}
    
\begin{document}

\title{VisAgent: Narrative-Preserving Story Visualization Framework\\

\thanks{*: Equal contribution, $\dagger$: Corresponding author}
}

%

\author{\IEEEauthorblockN{Seungkwon Kim\textsuperscript{*}}
\IEEEauthorblockA{\textit{NAVER WEBTOON AI} \\
Republic of Korea}
\and
\IEEEauthorblockN{GyuTae Park\textsuperscript{*}}
\IEEEauthorblockA{\textit{NAVER WEBTOON AI} \\
\textit{Seoul National University}\\
Republic of Korea}
\and
\IEEEauthorblockN{Sangyeon Kim}
\IEEEauthorblockA{\textit{NAVER WEBTOON AI} \\
Republic of Korea}
\and
\IEEEauthorblockN{Seung-Hun Nam\textsuperscript{$\dagger$}}
\IEEEauthorblockA{\textit{NAVER WEBTOON AI} \\
Republic of Korea}
}

\maketitle

\begin{abstract}
Story visualization is the transformation of narrative elements into image sequences.
While existing research has primarily focused on visual contextual coherence, the deeper narrative essence of stories often remains overlooked.
This limitation hinders the practical application of these approaches, as generated images frequently fail to capture the intended meaning and nuances of the narrative fully.
To address these challenges, we propose \textbf{VisAgent}, a training-free multi-agent framework designed to comprehend and visualize pivotal scenes within a given story.
By considering story distillation, semantic consistency, and contextual coherence, VisAgent employs an agentic workflow. In this workflow, multiple specialized agents collaborate to: (i) refine layered prompts based on the narrative structure and (ii) seamlessly integrate \gt{generated} elements, including refined prompts, scene elements, and subject placement, into the final image.
The empirically validated effectiveness confirms the framework's suitability for practical story visualization applications.
\end{abstract}


\section{Introduction}
\label{sec:intro}


Story visualization, the process of translating narrative content into visual representations, presents a significant challenge due to the need to convey the story’s essence to a diverse audience~\cite{segel2010narrative,song2020character}.
Recent advancements in diffusion models (DMs), such as Stable Diffusion (SD)~\cite{rombach2022high}, and large language models (LLMs) like GPT-4~\cite{openai2023gpt4}, have significantly enhanced story visualization capabilities.
DMs facilitate the generation of coherent and high-quality scenes by effectively capturing complex conditional distributions.
Meanwhile, LLMs complement this process by providing accurate text-to-visual alignment\textemdash termed \textit{semantic consistency}, ensuring that generated images closely reflect the narrative context and character properties, such as appearance, actions, and interactions~\cite{kim2024framework,guo2024large,zhang2023adding}.


In the developing field of story visualization, early efforts were centered around generating isolated scenes utilizing DMs and manual prompts for text-to-image (T2I) generation~\cite{rombach2022high,podell2023sdxl}.
As the field progressed, the focus shifted towards more sophisticated multi-scene visualizations with training emphasizing character identity preservation\textemdash \textit{contextual consistency}~\cite{hu2021lora,wang2024instantid,li2024photomaker,wang2023autostory,gong2023talecrafter}.
For effective multi-scene generation, precise foreground (FG) element placement within each background (BG) is crucial.
Recent advancements in layout suggestion techniques, driven by language models with vision capabilities~\cite{li2023gligen,lian2023llmgrounded,liu2024intelligent}, have introduced innovative training-free approaches that enhance flexibility, implementation, and computational efficiency in visual storytelling~\cite{zhou2024storydiffusion,cheng2024theatergen,cheng2024autostudio}.
Beyond fixed sequential prompts and story completion concepts~\cite{pan2024synthesizing,yang2024seed}, emerging approaches focus on \textit{story distillation}, a process that refines a given plain narrative into well-crafted prompts for multi-scene visualization~\cite{wang2023autostory,gong2023talecrafter}.

\begin{figure}[t]
\centering{\includegraphics[width=0.99\linewidth]{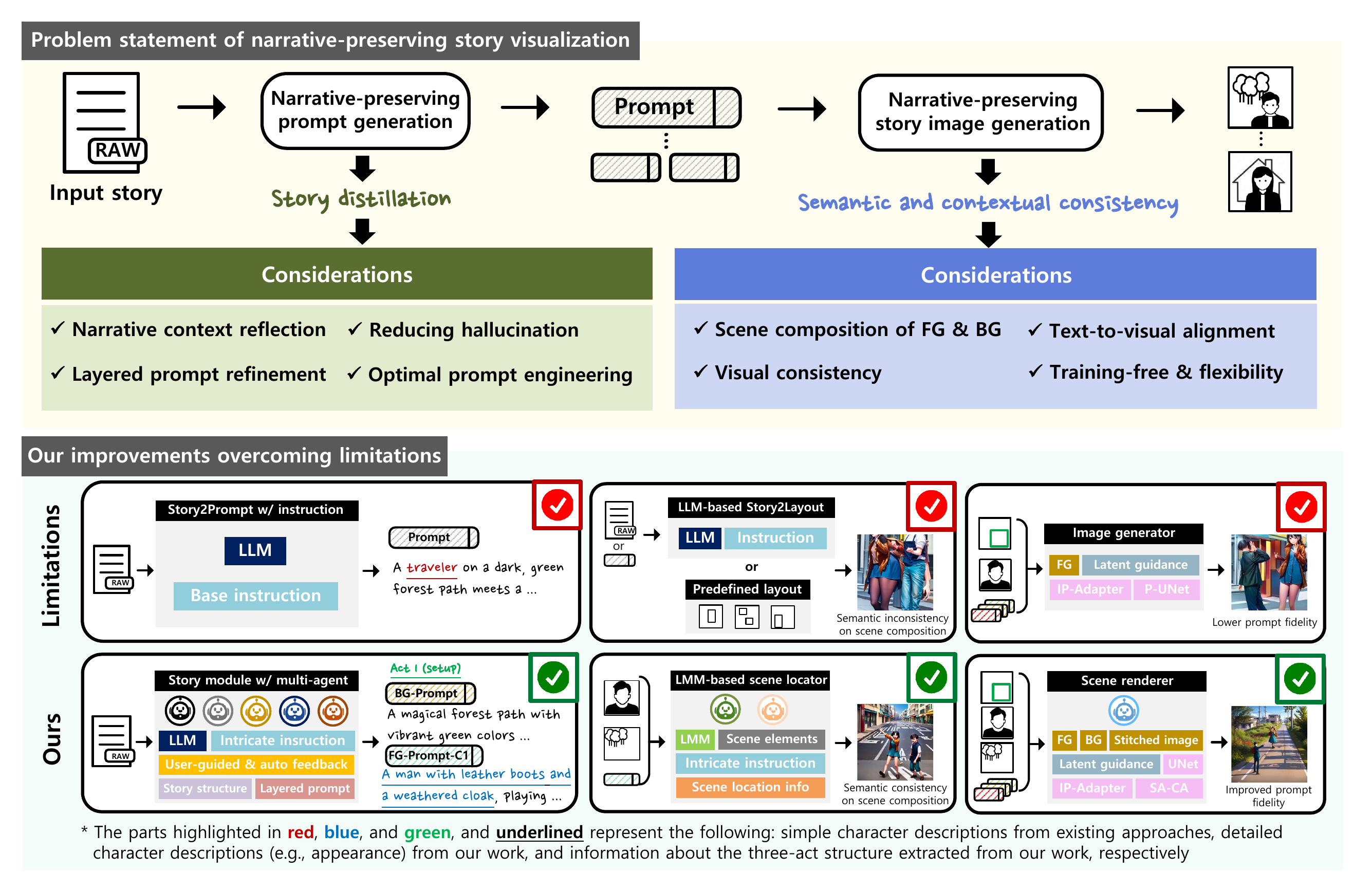}}
\vspace{-6mm}
\caption{Problem statement of narrative-preserving story visualization and our improvements for overcoming limitations.}
\vspace{-6mm}
\label{fig_overview}
\end{figure}


Through the analysis of existing literature, we have identified and separated two core components essential for narrative-preserving story visualization (Fig.~\ref{fig_overview}).
The primary objectives of these components are story distillation, semantic consistency, and contextual consistency.
Consideration in story distillation involves refining the detailed information of the scenes, including background (BG) and foreground (FG), into prompts suitable for diffusion models (DMs) to interpret, while maintaining the narrative context of the original story without inducing hallucinations.
When generating multiple scenes with prompts, it is crucial to consider the compositional and stylistic harmony between FG and BG elements, while maintaining the benefits of a training-free approach and ensuring coherence across multiple scenes.
Existing approaches often face limitations, as illustrated in the lower part of Fig.~\ref{fig_overview}.

\begin{figure*}[t]
\centering{\includegraphics[width=0.99\linewidth]{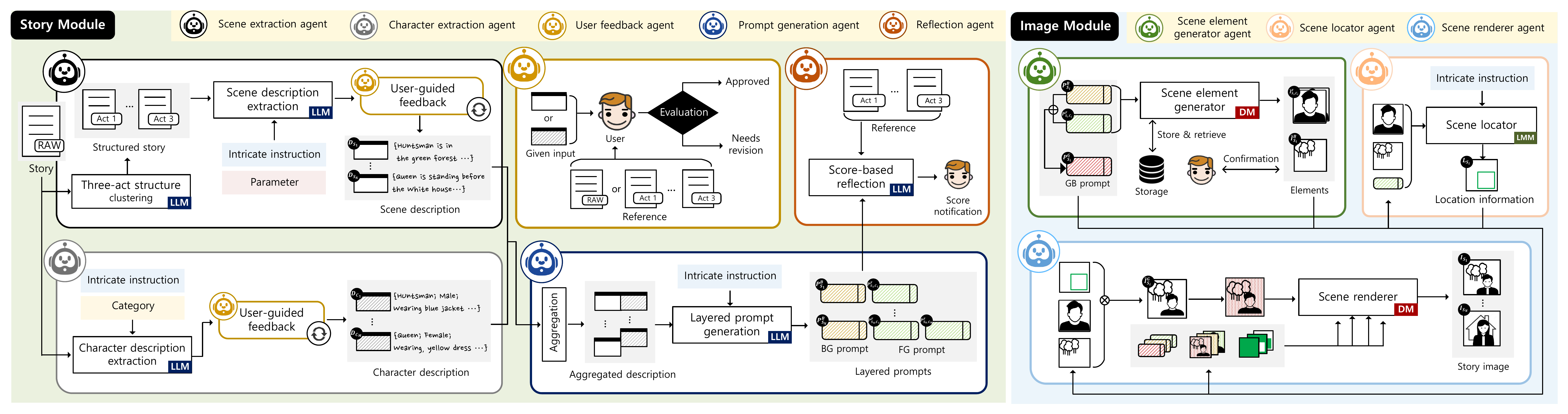}}
\vspace{-3mm}
\caption{Architectural overview of VisAgent, a multi-agent framework designed for story visualization. The recycle symbol (\faRefresh) represents a process that repeats until approval. $\oplus$ refers to the process of generating a GB prompt by concatenating BG and FG prompts to each scene while $\otimes$ specifically denotes a segmentation-based image stitching process.}
\vspace{-4mm}
\label{fig_VisAgent_framework}
\end{figure*}

To overcome these limitations, we propose a multi-agent framework named \textbf{VisAgent} to enable narrative-preserving story visualization with exquisite perceptual quality.
VisAgent comprises the following two core modules: \textit{story module} and \textit{image module}.
The story module leverages LLM-based agents to comprehend the input story in a three-act structure format, recognize the narrative context, and refine the extracted context information into well-crafted layered prompts (e.g., FG and BG prompts).
The resultant prompts are optimized for seamless interpretation by DMs, considering prompt engineering principles.
In addition, to minimize hallucinations and ensure rationale verification, the framework incorporates user-guided and automated agents for feedback and reflection on intermediate and final outputs. 
\gt{The image module, which is composed of three agents, is designed to generate narrative-preserving story images from layered prompts. To achieve this, it requires high fidelity in the FG and BG images, along with the appropriate placement of FG subjects within the BG to align with the narrative context.}
In the initial stage, the module prepares scene elements (e.g., character and BG images) in the scene element generator agent.
\gt{Subsequently, a scene locator agent, powered by a large multi-modal model (LMM), uses these elements to generate a layout for subject placement.}
Finally, the scene renderer agent integrates all element images and the location information to generate the final rendered image using the proposed semantic-aware cross-attention layer. As emphasized in Fig.~\ref{fig_overview}, our contributions are as follows. 

\begin{itemize}[leftmargin=3mm]
\item \gt{We propose} a story module, a specialized multi-agent framework that refines a given story into layered prompts by analyzing its narrative structure and distilling key events and character attributes, \gt{achieving effective} story distillation.

\item \gt{We propose an image module comprising multiple agents that collaboratively generate narrative-consistent story images by separately generating FG and BG, determining subject placement, and rendering the final image while ensuring semantic coherence with a novel semantic-aware cross-attention layer.} 

\item \nam{Our framework, designed with the interplay of multi-agents and user feedback, \gt{provides} an application that enables users to efficiently visualize their own stories, realizing their creative idea.}
\end{itemize}
\color{black}

\section{Proposed Framework}
\label{sec:proposed_framework}

Our framework, VisAgent, comprises story and image modules.
All agents within the framework are meticulously devised to collaborate, optimizing a multi-modal setting for story visualization.

\subsection{Preliminaries}

Let $R$ denote the input story in plain text format, with $N$ representing the number of scenes for story distillation and $M$ denoting the number of characters featured in $R$.
In this study, the elements corresponding to each scene and character are denoted by \(S_i\) and \(C_j\), where \(i \in \{0, 1, \dots, N-1\}\) and \(j \in \{0, 1, \dots, M-1\}\).
We define \(D_{S_i}\) as the scene description, \(D_{C_j}\) as the character description, \(P^{B}_{S_i}\) as the BG prompt, \(P^{F}_{S_i,C_j}\) as the FG prompt for a specific character, and \(P^{G}_{S_i}\) as the global (GB) prompt.
The BG and FG character images denoted as $\tilde{I}^{B}_{S_i}$ and $\tilde{I}^{F}_{S_i,C_j}$ respectively, are generated from \(P^{B}_{S_i}\) and \(P^{F}_{S_i,C_j}\).
The location of the subject, defined by the coordinates \((x_{\text{min}}, y_{\text{min}}, x_{\text{max}}, y_{\text{max}})\), is represented as \(L_{S_i}\), while the generated story image is denoted as \(I_{S_i}\).

\subsection{Story Module}
\label{sec:StoryModule}

\subsubsection{Narrative-Preserving Prompt Generation}

While LLMs have been extensively studied for their effectiveness in generating prompts tailored for DMs~\cite{lian2023llmgrounded, gong2023talecrafter,qin2024diffusiongpt,zhou2024storydiffusion}, existing approaches in story visualization often prioritize image quality over the preservation of the narrative structure~\cite{lian2023llmgrounded, liu2024intelligent}. 
This can result in generated images that fail to effectively convey the story’s intentions, potentially leading to incomplete or insufficient visualizations.

\subsubsection{Methodology}
\nam{To address this limitation, we develop the story module, a multi-agent framework that incorporates the human-in-the-loop concept and introduces a narrative-preserving concept for the first time in story visualization tasks.
Our module analyzes narrative structures from plain text input and refines them into layered prompts for BG and FG, enabling story distillation.
In this way, the refined output can preserve the key scenes of the story, while incorporating detailed descriptions for both BG and FG (see Table~\ref{table_qualitative_eval_test})
}

In detail, the story module is defined as $SM(R,\epsilon_{S},\delta,N)$, where \(\epsilon_{S}\) and \(\delta\) represent the devised intricate instruction and predefined category, respectively.
In detail, the input story $R$ is thoroughly analyzed and transformed into effective prompts for a generative model while preserving its essential narratives.
Our approach is grounded in the assumption that all stories, to varying degrees, rely on specific narrative structures to enhance their appeal.
Central to this approach is the \textit{scene extraction agent}, which deconstructs and distills the story utilizing the storytelling structure (i.e., three-act structure~\cite{papalampidi2020screenplay}), a widely recognized and long-standing framework in storytelling.
In addition, the \textit{character extraction agent} identifies and extracts descriptions about all characters, including their attire, gender, and other attributes defined by the predefined category.
If the attire is unspecified, the agent infers details from context, essential for defining character style in DM prompts.


Although these agents effectively identify crucial scenes and characters automatically, the \textit{user feedback agent} allows users to confirm or modify parts of the outputs (\(D_{S_i}\) and \(D_{C_j}\)), ensuring the preservation of all critical information and reducing hallucinations throughout the distillation process.
Based on the user-confirmed results, the \textit{prompt generation agent} then generates separate BG and FG prompts (\(P^{B}_{S_i}\) and \(P^{F}_{S_i,C_j}\)), each covering mutually exclusive aspects of the scene and having a format tailored to the targeted DM by applying prompt engineering principles.
Finally, in addition to the LLM’s capability for textual similarity assessment, the \textit{reflection agent} conducts a thorough review by comparing the prompts with the corresponding story segments, informing users of any potential deviations caused by the story module.
The resulting layered prompts offer two advantages: (i) preserving the essential narrative elements of the original story, and (ii) aligning with the standard graphic narrative synthesis process for FG and BG composition, which are crucial components for the image module.

\subsection{Image Module}
\label{sec:Image_Module}

\subsubsection{Narrative-Preserving Story Image Generation}

\gt{To generate a narrative-preserving story image, it is essential to create an image that achieves high prompt fidelity for both FG and BG elements and ensures that FG subjects are appropriately positioned and integrated within the BG. 
Inspired by a common process of creating cartoons, we generate FG and BG elements separately, determine the layout for placing FG elements on the BG, and finally produce the complete story image. This approach preserves the quality of each scene element while considering its global context, resulting in a narrative-preserving story image.}


\begin{figure}[t]
\centering{\includegraphics[width=0.98\linewidth]{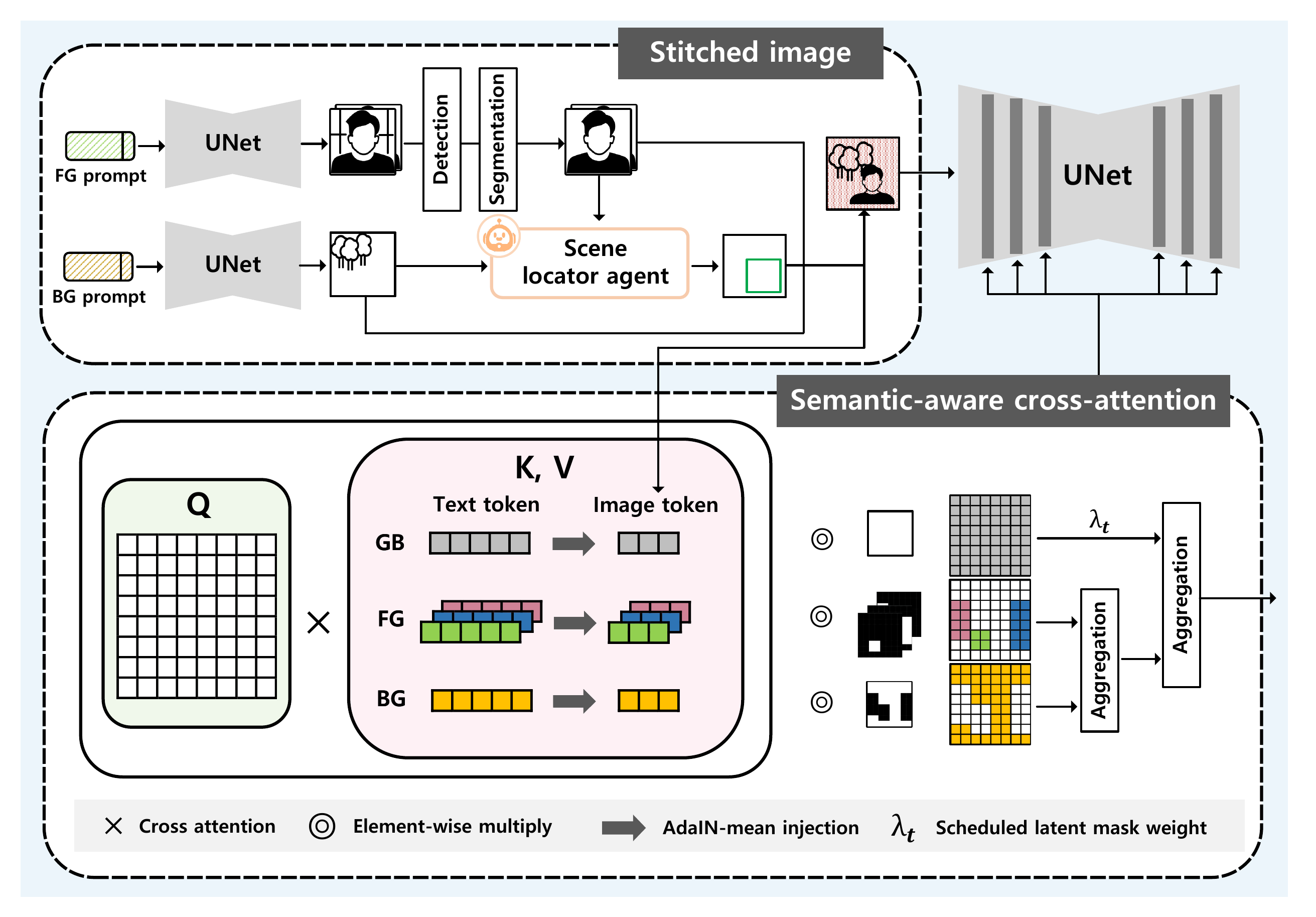}}
\vspace{-3mm}
\caption{Schematic of semantic-aware cross-attention layer.}
\vspace{-4mm}
\label{fig_SACA}
\end{figure}

\subsubsection{Methodology}

From this perspective, we propose the image module, composed of three specifically designed agents to generate story scene images (see Fig~\ref{fig_VisAgent_framework}).
The image module is defined as $IM(P^{B}_{S_i},P^{F}_{S_i,C_j},\epsilon_{I})$, where \(\epsilon_{I}\) represents the devised intricate instruction.

\gt{First}, the \textit{scene element generator agent} generates each element image (i.e., FG character images and a BG image) from refined layered prompts, which serve as key components in the composition of resultant images.
Note that we adopt a strategy of utilizing an IP-Adapter~\cite{ye2023ip-adapter} and a subject storage, similar to~\cite{cheng2024autostudio, cheng2024theatergen}, to manage consistency between element images (i.e., contextual consistency).
Users may optionally repeat and confirm this process multiple times until the scene elements are satisfactorily produced.

\gt{Subsequently, the \textit{scene locator agent}, powered by an LMM model, suggests suitable FG placement within the BG, considering the semantic context of layered prompts and the BG image and ensuring a harmonious scene composition.}
Unlike existing approaches \cite{cheng2024autostudio, cheng2024theatergen} that solely rely on prompts for scene layout determination, our agent leverages visual context from the scene element images to ensure accurate scene composition, thereby significantly enhancing semantic consistency and achieving narrative-preserving scene synthesis.

\gt{Finally}, the \textit{scene renderer agent} generates a story scene image \gt{using} a stitched image as latent guidance input, along with the scene renderer, a DM that incorporates a modified version of the cross-attention layer.
Initially, the agent generates a stitched image by meticulously integrating the FG with the BG, employing an open-vocabulary detection model~\cite{liu2023grounding} and a segmentation model~\cite{kirillov2023segany} in a layer-stacking manner.
Subsequently, the stitched image is encoded and passed through the forward diffusion process, and then used as a latent guidance for the input of the scene renderer.
In particular, to augment semantic consistency, we propose a novel \gt{Semantic-Aware Cross-Attention (SA-CA)} layer that comprehensively utilizes all layered prompts, scene elements (FG and BG), and latent guidance inputs (see Fig.~\ref{fig_SACA}).
Although it is inspired by the parallel text and cross-attention mechanisms of P-UNet~\cite{cheng2024autostudio}, the following key differences are introduced to further enhance semantic consistency.


\begin{itemize}[leftmargin=3mm]


\item \gt{\textit{Using BG and stitched image as reference and guidance input}: Our baseline \cite{cheng2024autostudio} employs FG images as reference image tokens in the modified cross-attention layer. The SA-CA layer extends this by leveraging FG, BG, and a stitched image as reference tokens for their respective regions, to generate a semantically coherent scene. Besides, we use the stitched image as latent guidance instead of a black BG with the pasted FG image. As FG and BG are generated independently earlier, the SA-CA layer is expected to enhance prompt fidelity over the baseline, particularly for the BG.} 


\item \gt{\textit{Global latent aggregation for global region}: Cross-attention is applied separately to the FG and BG regions using their respective layered text and image prompt tokens, as well as across the entire region using a global text (combining FG and BG) prompt token and a global image prompt token derived from the stitched image. To aggregate all latents, we introduce \(\lambda_{t}\) adjusting the influence of a global latent during scene generation. We adopt a stepwise strategy where \(\lambda_{t}\) progressively increases over the timesteps.} 



\item \gt{\textit{Token alignment between text and image prompt token}: Inspired by \cite{wu2024infinite}, we apply an adaptive mean normalization (AdaIN-mean) operation between text and image prompt tokens to enhance semantic consistency while maintaining identity fidelity, as shown by the thick gray arrow in Fig.~\ref{fig_SACA}. For FG, BG and GB latent, the key and value from the image tokens are aligned with those from the text tokens before applying cross-attention.}

\end{itemize}

\begin{table}[!t]
\caption{Results on narrative-preserving prompt generation} 
\vspace{-2mm}
    \resizebox{\linewidth}{!}{
    \centering
    \setlength{\tabcolsep}{3.5pt}
    \begin{tabular}{c c c}
    \hline
    \hline
    Method & Scene & Refined prompt \\
    \hline
    & \multirow{1}{*}{\shortstack{\#1}} & \multicolumn{1}{l}{\textit{A young boy, Jack, standing at the base of a towering beanstalk reaching into a mystical sky, surrounded by lush, vibrant greenery}}\\
    \cmidrule(r){2-3}
    GPT-4o & \#2 & \multicolumn{1}{l}{\textit{A massive, golden giant's castle in the clouds, with Jack peeking in through a grand window, his face full of awe and trepidation}}\\
    \cmidrule(r){2-3}
    w/ base & \#3 & \multicolumn{1}{l}{\textit{Jack climbing a colossal, twisting beanstalk against rolling clouds and a shimmering sunset, filled with adventure and wonder}} \\
    \cmidrule(r){2-3}
    ins.~\cite{gong2023talecrafter}& \#4 & \multicolumn{1}{l}{\textit{An enchanted giant’s treasure room filled with golden coins, shimmering jewels, and magical artifacts, with Jack cautiously exploring}}\\
    \cmidrule(r){2-3}
    &  \#5 & \multicolumn{1}{l}{\textit{A dramatic moment of Jack slaying a fierce giant, with intense action and dynamic lighting highlighting the epic confrontation}}\\
    \hline
    &  & \multicolumn{1}{l}{\textit{\textbf{BG}: A humble rusty and weathered traditional market, no building, surrounded by a few sparse trees and a patchy garden, highres}} \\
    & \#1: & \multicolumn{1}{l}{\textit{\hspace{0.5cm} detailed, soft lighting, daytime}} \\
    & Act 1 & \multicolumn{1}{l}{\textit{\textbf{FG-C1}: A small boy with worn-out blue medieval clothing, standing, handing over a cow}} \\
    & & \multicolumn{1}{l}{\textit{\textbf{FG-C2}: A old man with worn-out medieval merchant clothing, standing, holding a basket of magical beans}}\\
    \cmidrule(r){2-3}
    & \#2: & \multicolumn{1}{l}{\textit{\textbf{BG}: A towering, fantastically gigantic beanstalk spiraling up into the sky, disappearing into the clouds, a blue sky with wisps of}}\\
    & Act 1 & \multicolumn{1}{l}{\textit{\hspace{0.5cm} clouds, highres, detailed, soft lighting, daytime}}\\
    &  & \multicolumn{1}{l}{\textit{\textbf{FG-C1}: A small boy with worn-out blue medieval clothing, climbing, holding onto the gigantic beanstalk}}\\
    \cmidrule(r){2-3}
    Story& \#3: & \multicolumn{1}{l}{\textit{\textbf{BG}: An old medieval cottage surrounded by trees, clear sky, highres, detailed, soft lighting, daytime}}\\
    module& Act 2 & \multicolumn{1}{l}{\textit{\textbf{FG-C1}:A small boy with worn-out blue medieval clothing, standing, discovering a mysterious cottage}}\\
    \cmidrule(r){2-3}
    & \#4: & \multicolumn{1}{l}{\textit{\textbf{BG}: An antique interior of mysterious medieval cottage, furniture, beams, highres, detailed, soft lighting, daytime}}\\
    & Act 2 & \multicolumn{1}{l}{\textit{\textbf{FG-C1}: A small boy with worn-out blue medieval clothing, standing, holding a stolen big white goose, looking scared}} \\
    & & \multicolumn{1}{l}{\textit{\textbf{FG-C3}: A giant human monster with muscle, beard, big nose and with black village clothing, sitting, closed eyes}}\\
    \cmidrule(r){2-3}
    & \#5: & \multicolumn{1}{l}{\textit{\textbf{BG}: A towering, fantastically gigantic chunk of beanstalk, meadow, grass, clouds, trees, highres, detailed, soft lighting, daytime}}\\
    & Act 3 & \multicolumn{1}{l}{\textit{\textbf{FG-C1}: \color{black} A small boy with worn-out blue medieval clothing, hiding, standing next to a gigantic beanstalk} } \\
    & & \multicolumn{1}{l}{\textit{\textbf{FG-C3}: \color{black} A giant human monster with muscle, beard, big nose and with black village clothing\color{black}, falling from sky, floating in the air}}\\
    \hline
    \hline
    \end{tabular}
    }   
    \begin{tablenotes}
    \tiny
    \item * Here, Acts 1, 2, and 3 correspond to a three-act structure’s setup, conflict, and resolution. C1 to C3 each represent the index of a character. In the story module, scenes such as exchanging the cow for beans (\#1) and stealing treasures (\#3) are highlighted because of their critical narrative significance. In contrast, the baseline focuses merely on major events, often disconnected from the overall narrative. 
    \end{tablenotes}
    \vspace{-1mm}
    \label{table_qualitative_eval_test}
\end{table}

\begin{figure}[t]
\centering{\includegraphics[width=0.98\linewidth]{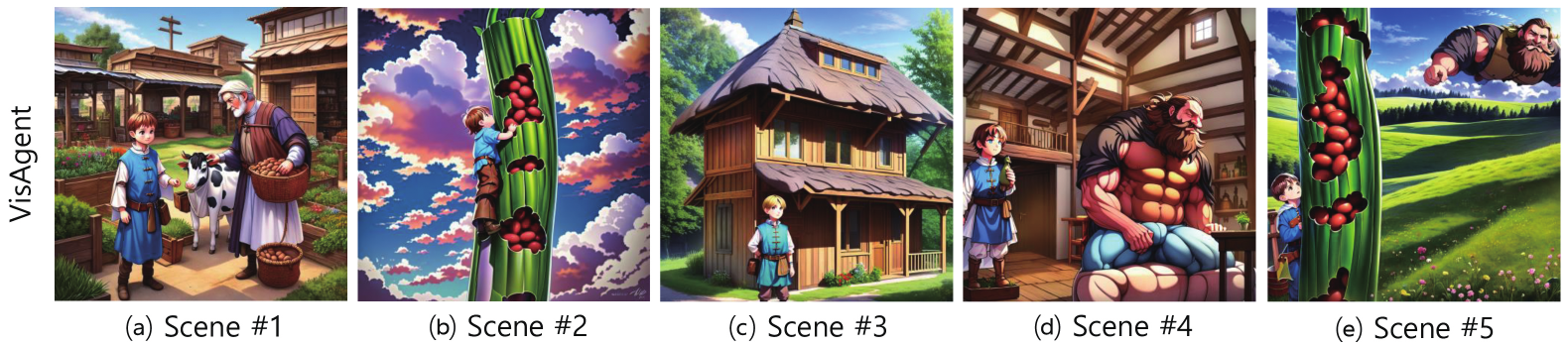}}
\vspace{-1mm}
\caption{Results of qualitative evaluation: narrative-preserving story visualization results of VisAgent using refined layered prompts as listed in Table~\ref{table_qualitative_eval_test}.}
\vspace{-2mm}
\label{fig_qualitative_evaluation_narrative_preserving}
\end{figure}

\begin{figure}[t]
\centering{\includegraphics[width=0.98\linewidth]{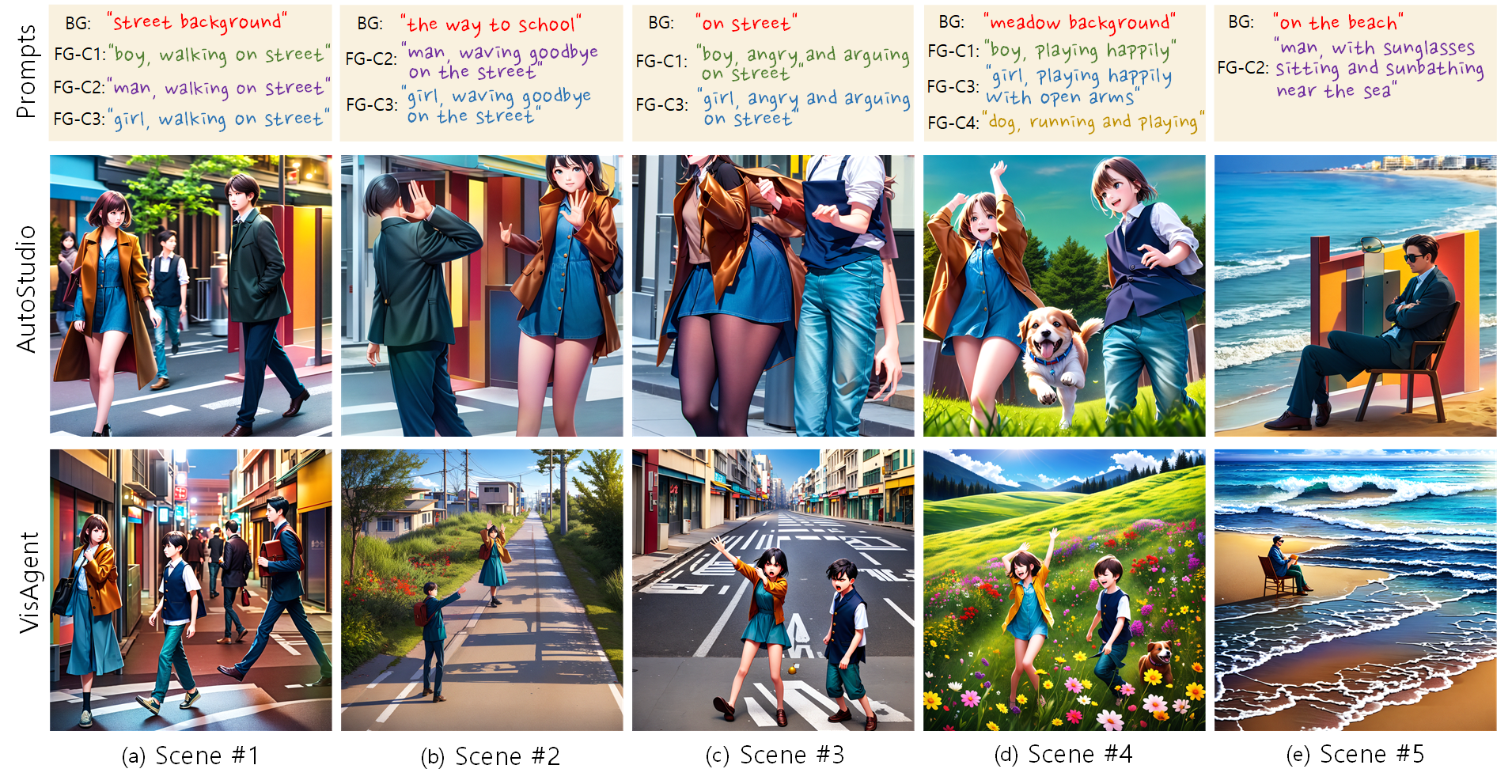}}
\vspace{-2mm}
\caption{Results of qualitative evaluation: performance analysis of five example prompts-based story visualizations compared to the baseline.}
\vspace{-1mm}
\label{fig_qualitative_evaluation_normal_prompt}
\end{figure}


\begin{figure}[t]
\centering{\includegraphics[width=0.98\linewidth]{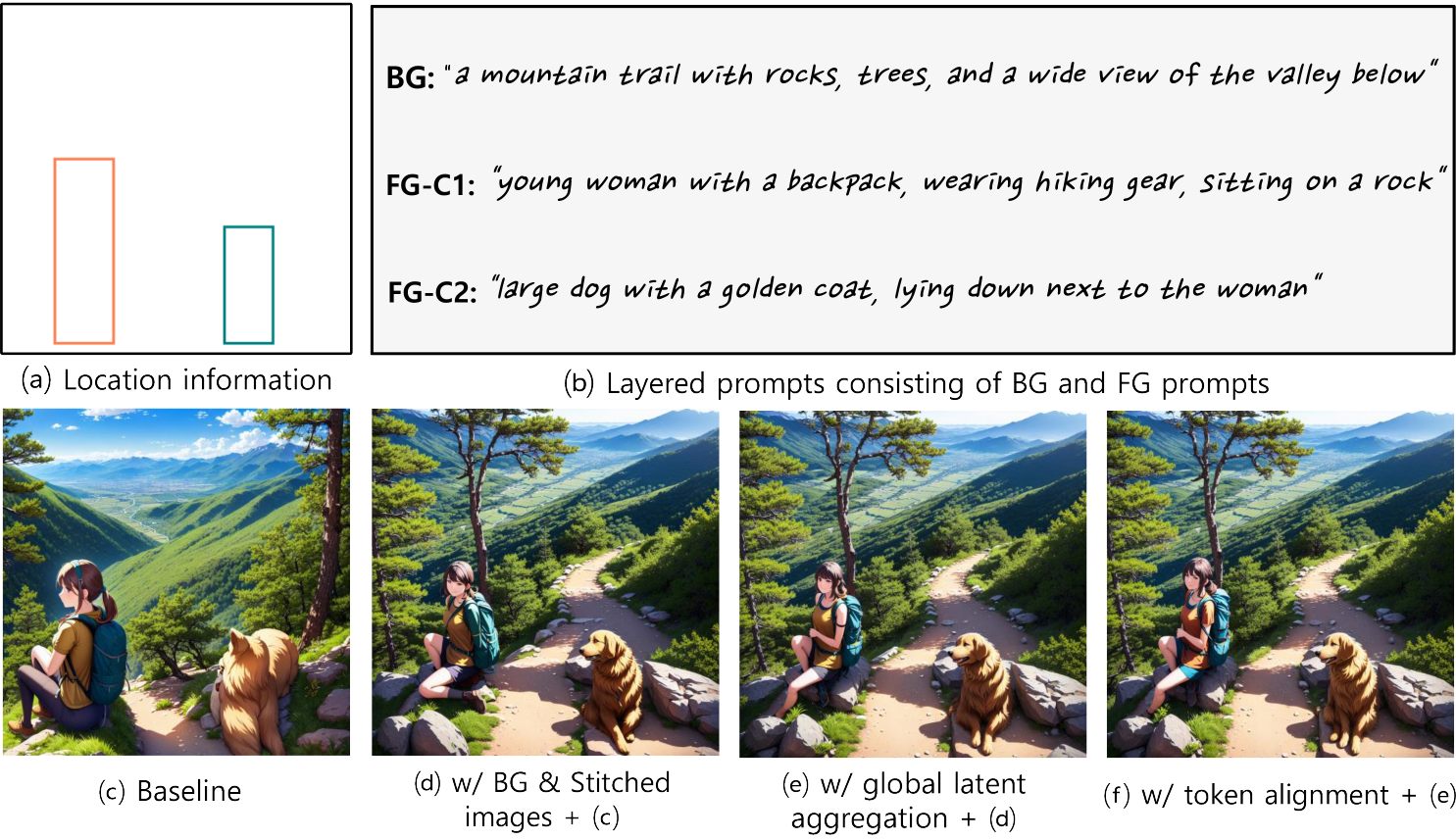}}
\vspace{-2mm}
\caption{Results of ablation study on SA-CA of image module.}
\vspace{-3mm}
\label{fig_ablation_study}
\end{figure}

\section{Experiments}

\subsection{Settings}

\subsubsection{Implementation Details}

We adopt GPT-4o, a variant based on GPT-4~\cite{openai2023gpt4}, as LLM and LMM. The multi-agent system in the story module is implemented based on the \texttt{LangGraph} framework~\cite{langgraph}.
For the story module, plain text about \textit{$<$Jack and the Beanstalk$>$} is employed as input story\footnote{https://americanliterature.com/childrens-stories/jack-and-the-beanstalk}, featuring three characters and narrative key moments, and the $N$ for story distillation is set to 5.
For the image module, we adopt a model based on SD v1.5, as specified in~\cite{cheng2024autostudio}, \gt{using a DDIM sampler with 30 steps.}
\(\lambda_{t}\) is set to 0.1 for Steps 1--10, 0.3 for Steps 11--20, and 0.5 for Steps 21--30, incrementing at intervals of ten steps. The evaluations are completed with one NVIDIA A100 GPU with 80 GB of GPU memory.

\subsubsection{\gt{Evaluations}}

The story and image modules constituting the VisAgent are evaluated via quantitative and qualitative evaluations. Referencing~\cite{hessel2021clipscore,cheng2024theatergen}, we select the quantitative metrics of Fréchet inception distance (FID) and character–character similarity (CCS) to evaluate contextual consistency and text-image similarity (TIS) to assess semantic consistency.
\gt{To quantitatively evaluate the scene renderer agent, a new VisAgent benchmark is introduced, comprising 400 narrative stories generated by an LLM. Based on a format similar to CMIGBench~\cite{cheng2024theatergen}, it includes richer FG and BG prompts to assess the ability to generate detailed and narrative scene content.}
In addition, we exhibit the results for qualitative evaluation of the distilled story and the visualized scene image, and conduct a human preference study via an A/B test with 20 volunteers.


\subsubsection{Baselines}
To evaluate the story distillation capabilities of our story module, we compared its performance to the baseline approach (i.e., GPT-4o with base instruction) specified in~\cite{gong2023talecrafter}.
We compare our image module with the state-of-the-art training-free method, AutoStudio~\cite{cheng2024autostudio}, from the perspective of visualization performance.

\subsection{Experimental Results}

\subsubsection{Results of Story Module}

Our story module effectively captures key narrative elements, including build-up scenes leading to the climax and connecting scenes between major events (see Table~\ref{table_qualitative_eval_test}).
In contrast, the baseline approach often produces an irregular distribution of scenes, neglecting narrative flow and focusing solely on major events.
Our module also ensures consistent character styles (e.g., appearance) across prompts and provides refined layered prompts, enhancing fidelity and narrative representation in the visualization process.
User evaluations, assessing the distillation quality of the prompts listed in Table~\ref{table_qualitative_eval_test} and their suitability for DMs, further confirm the effectiveness of our module (see Table~\ref{table_human_evaluation}).

\subsubsection{Results of Image Module}

First, we perform a qualitative evaluation on visualized resultant images.
As shown in Fig.~\ref{fig_qualitative_evaluation_narrative_preserving}, the image module achieves narrative-preserving visualization with layered prompts refined from the plain story. In a comparison with \cite{cheng2024autostudio}, our module demonstrated superior quality across five example scenes as illustrated in Fig.\ref{fig_qualitative_evaluation_normal_prompt}). This achievement stems from leveraging our scene locator agent, which considers semantic composition and other agents enhancing semantic consistency and contextual consistency with the layered prompts. 

\gt{Second, Table~\ref{table_quantitative_eval} presents the quantitative results of the scene renderer agent which has a SA-CA layer, compared to the baseline, using FID, CCS, and TIS metrics, demonstrating improved performance across both benchmarks. Note that the scene locator is not used in this experiment to ensure a fair comparison, and the metrics of the baseline were reproduced using the official code in our setting.} 

\sy{Moreover, Fig.~\ref{fig_ablation_study} presents output images illustrating the application of a several strategies to the baseline, resulting in a SA-CA layer. Specifically, Fig.~\ref{fig_ablation_study}(a)-(b) depict the input components, while Fig.~\ref{fig_ablation_study}(d)-(f) show output images generated by the SA-CA with submodules specified in the subcaptions.
Compared to the baseline, Fig.~\ref{fig_ablation_study}(d) demonstrates improved prompt fidelity for the BG prompt, \textit{''a mountain trail,''} and produces a more natural image of FG-C2. By integrating the global latent with $\lambda_{t}$, Fig.~\ref{fig_ablation_study}(e) shows contextually improved image, such as FG-C1's \textit{``sitting on a rock,''} with a more natural pose. As a final step, Fig.~\ref{fig_ablation_study}(f) demonstrates slightly improved quality through the incorporation of token alignment.} 

Finally, we conduct a human preference study to provide an intuitive evaluation of our work compared to the baseline~\cite{cheng2024autostudio}. To this end, we generated story scene images with prompts and exposed them with guidelines (i.e., the criteria for measuring contextual consistency include prompt fidelity and the scene composition between the BG and FG), requesting volunteers to perform an A/B test. The results presented in Table~\ref{table_human_evaluation} indicate the superiority of the proposed module.


\begin{table}[!t]
\caption{Results of quantitative evaluation} 
\vspace{-2mm}
\resizebox{\linewidth}{!}{
    \centering
    \scriptsize
    \begin{tabular}{c c c c c c c}
    \hline
    \hline
    \multirow{2}{*}{\shortstack{Method}} & \multicolumn{3}{c}{VisAgent Benchmark} & \multicolumn{3}{c}{CMIGBench} \\
    \cmidrule(r){2-4} \cmidrule(r){5-7}
    & TIS (\%) $\uparrow$ &FID $ \downarrow$ & CCS (\%) $\uparrow$ & TIS (\%) $\uparrow$ & FID $\downarrow$ & CCS (\%) $\uparrow$ \\
    \hline
    AutoStudio & 25.04 & 267.48 & 83.42 & 31.90\textsuperscript{$\dagger$} & 246.85\textsuperscript{$\dagger$} & 79.02\textsuperscript{$\dagger$} \\
    VisAgent  & \textbf{25.43} & \textbf{263.74} & \textbf{83.76} & \textbf{32.58} & \textbf{243.11} & \textbf{80.14} \\
    \hline
    \hline
    \end{tabular}
    \vspace{-2mm}
    }
    \begin{tablenotes}
    \footnotesize
    \item $\dagger$ denotes results reproduced by our implementation using the official code.
    \end{tablenotes}
    \label{table_quantitative_eval}
\end{table}

\begin{table}[!t]
\caption{Results of human evaluation (\%)} 
\vspace{-2mm}
    \resizebox{\linewidth}{!}{
    \centering
    \centering
    \begin{tabular}{c c c c c}
    \hline
    \hline
    \multirow{2}{*}{\shortstack{Metric}} & \multicolumn{2}{c}{Story distillation} & \multicolumn{2}{c}{Semantic and contextual consistency} \\ 
    \cmidrule(r){2-3} \cmidrule(r){4-5}
    & GPT-4o w/ \cite{gong2023talecrafter} & VisAgent & AutoStudio~\cite{cheng2024autostudio} & VisAgent  \\
    \hline
    User score& 17.78 & \textbf{82.22} & 15.56 & \textbf{84.44}\\
    \hline
    \hline
    \end{tabular}
    \label{table_human_evaluation}
    }
\vspace{-2mm}
\end{table}

\section{Conclusion}
This study introduced VisAgent, a multi-agent framework designed for narrative-preserving story visualization.
The framework’s story module effectively refines the input story through a sophisticated multi-agentic workflow, producing layered prompts that accurately capture the narrative context while minimizing hallucinations.
\gt{The image module, composed of scene element generator, scene locator and scene renderer agents, generate narrative-preserving story image by effectively integrating scene elements}. Experimental results validate the effectiveness of VisAgent.

\vfill\pagebreak

\bibliographystyle{IEEEtran}

\end{document}